\documentclass[a4paper, fleqn]{cas-dc}
\usepackage{bm}

\usepackage{amsthm}

\usepackage{graphicx}
\usepackage{svg}
\usepackage{epstopdf}
\usepackage{caption}
\usepackage{subfigure}
\usepackage{float}
\usepackage[hypcap = true, hypcapspace = 2cm]{caption}
\usepackage{makecell}
\usepackage{bigstrut}
\usepackage{tabularx}
\usepackage{threeparttable} % 表格注释环境
 
\usepackage{amsmath}
\usepackage[ruled, linesnumbered]{algorithm2e}
\SetKwRepeat{Do}{do}{while}

\usepackage{hyperref} % 目前存在部分问题，需解决：1.参考文献超链接问题
\hypersetup{colorlinks=true,
            linkcolor=blue,   %修改此处为你想要的颜色
            anchorcolor=black, %修改此处为你想要的颜色
            citecolor=blue,    %设置文中引用参考文献的颜色
            urlcolor=blue      %设置DOI的颜色为蓝色
            }

\usepackage[numbers, sort&compress]{natbib}
\usepackage[numbers]{natbib}
\usepackage{appendix}
\renewcommand\appendix{\par
    \setcounter{section}{0}
    \setcounter{subsection}{0}
    \gdef\thesection{Appendix \Alph{section}}}

\def\tsc#1{\csdef{#1}{\textsc{\lowercase{#1}}\xspace}}
\tsc{WGM}
\tsc{QE}

\begin{document}
\let\WriteBookmarks\relax
\def\floatpagepagefraction{1}
\def\textpagefraction{.001}
\let\printorcid\relax % 去掉页面下方的ORCID

% 文章短标题：Short title 和 Short author 是显示在页眉和页脚的信息，如果不需要可以直接注释掉。
% \shorttitle{<short title of the paper for running head>} 
% \shorttitle{Test Title}    

% 文章短作者
% \shortauthors{<short author list for running head>}
\shortauthors{Zhaozhong Y and Liangjie Gou et al.}

% 文章标题
\title[mode = title]{Domain adaption and physical constrains transfer learning for shale gas production}  

% 题目脚注
% \tnotemark[<tnote number>] 
% \tnotemark[1,2]
% \tnotetext[1]{This document is the results of the research project funded by the National Science Foundation.}
% \tnotetext[2]{The second title footnote which is a longer text matter to fill through the whole text width and overflow into another line in the footnotes area of the first page.}

% 作者信息

% \fnmark[1] % 标注作者顺序，不常用
% \ead{} % 填写作者邮箱
% \ead[url]{} % 填写作者个人网站，不常用
% \credit{} % 标注作者贡献，不常用
\author[1]{Zhaozhong Yang}[style=chinese]
\author[1, 3]{Liangjie Gou}[style=chinese]
\author[1, 2, 3]{Chao Min}[style=chinese]
% 添加通讯作者的星号
\cormark[1]
\author[1]{Duo Yi}[style=chinese]
\author[1]{Xiaogang Li}[style=chinese]
\author[2,3]{Guoquan Wen}[style=chinese]

\address[1]{National Key Laboratory of Oil and Gas Reservoir Geology and Exploitation, Chengdu 610500, China}
\address[2]{School of Science, Southwest Petroleum University, Chengdu 610500, China}
\address[3]{Institute for Artificial Intelligence, Southwest Petroleum University, Chengdu 610500, China}

\cortext[1]{School of Science, Southwest Petroleum University, Chengdu 610500, China. Email address: minchao@swpu.edu.cn (Chao Min).} % 通讯作者信息

% 摘要
\begin{abstract}
Effective prediction of shale gas production is crucial for strategic reservoir development.  However, in new shale gas blocks, two main challenges are encountered: (1) the occurrence of negative transfer due to insufficient data, and (2) the limited interpretability of deep learning (DL) models. To tackle these problems, we propose a novel transfer learning methodology that utilizes domain adaptation and physical constraints. This methodology effectively employs historical data from the source domain to reduce negative transfer from the data distribution perspective, while also using physical constraints to build a robust and reliable prediction model that integrates various types of data. The methodology starts by dividing the production data from the source domain into multiple subdomains, thereby enhancing data diversity. It then uses Maximum Mean Discrepancy (MMD) and global average distance measures to decide on the feasibility of transfer. Through domain adaptation, we integrate all transferable knowledge, resulting in a more comprehensive target model. Lastly, by incorporating drilling, completion, and geological data as physical constraints, we develop a hybrid model. This model, a combination of a multi-layer perceptron (MLP) and a Transformer (Transformer-MLP), is designed to maximize interpretability. Experimental validation in China's southwestern region confirms the method's effectiveness.
\end{abstract}

% 关键词:Each keyword is seperated by \sep

\maketitle % 编译前面的标题、作者与日期等信息, 不可缺少

% 文章正文文本
% 引言
\section{Introduction}
Shale gas, acclaimed for its relatively low environmental impact compared to other fossil fuels, is recognized as a practical avenue for achieving decarbonization \cite{a1hu2023greena1}. Accurate production prediction plays a vital role in evaluating reservoir development and formulating secondary stimulated strategies \cite{a2lee2019prediction02,a3li2020new03,a4niu2022development04,a5zhai2022prediction05}. It directly determines whether the stimulation measures are successful enough to meet the economic requirements \cite{a6wang2020modeling06}.

Production prediction is a highly anticipated research. To attain this challenging goal, extensive research has been conducted, primarily categorized into methods based on decline curve analysis (DCA) \cite{a7yehia2023comprehensive07}, numerical simulation \cite{a8jiang2023review08} and deep learning (DL) \cite{a9he2023shale09}. DCA is an empirical method that involves fitting historical production curves. The most widely used methods include the Arps decline method \cite{a10pratama2024probabilistic10}, the Valko stretch exponential production decline (SEPD) \cite{a11huang2023review11}, and the Duong rate decline for fractured-dominated fracture flow (Duong method) \cite{a12liu11process12}. DCA is straightforward but highly influenced by subjective factors, resulting in overly idealized result \cite{a13syed2022smart13}. Numerical simulation methods are employed to simulate complex shale gas flow mechanisms, providing a partial solution to the limitations of DCA. Lin et al. \cite{a14lin2023prediction14} studied the impact of fracture length and initial gas pressure on shale gas production by exploring multiple gas transport mechanisms in nanoscale pores and the variation of the permeability of the shale gas reservoir with nanoscale pore pressure. Yang et al. \cite{a15yang2023recent15} used digital rock core technology to investigate gas transport on the pore scale, considering factors such as desorption, surface diffusion, and sliding during shale gas production. However, these methods have limitations in practical applications. Accurate numerical simulation results often require a substantial investment in computational resources and time. Furthermore, the randomness and volatility of data can significantly impact predictive outcomes \cite{a16li2023prediction16}. 

In recent years, DL has gained substantial attention in production prediction \cite{a8jiang2023review08}. Viet et al. \cite{a54nguyen2022artificial54} presented a model for predicting cumulative shale gas production by combining Arps decline and artificial neural network, and verified the results based on numerical simulation data. Guo et al. \cite{a17guo2021prediction17} proposed a novel approach for predicting production from well logging data using DL, combining convolutional autoencoders and LSTM networks. Yang et al. \cite{a18yang2022physics18} developed a data-driven framework that employs physically constrained neural networks for production prediction, aiming to incorporate well-logging data and static geological parameters into LSTM to create a physically constrained neural network. FJ López-Flores et al. \cite{a26lopez2023development26} developed a DL model based on MLP, utilizing the Eagle Ford Formation study area dataset to predict cumulative shale gas production for 12 months and 90 days of return of water production. Qin et al. \cite{a19qin2023combined19} presented a model based on gated recurrent units and a multilayer perceptron (GRU-MLP) for prediction of horizontal well production of multistage hydraulic fracturing. They integrated the non-dominated sorting genetic algorithm II (NSGA II) into the model for automatic structural optimization. Li et al. \cite{a20li2022predicting20} proposed a Temporal Convolutional Network (TCN) that predicts shale gas production by capturing dependencies between the pressure of the previous wellhead and production. Kocoglu et al. \cite{a55kocoglu2021application55} introduced a Bi-LSTM model that effectively handles sequential data to explore the relationship between explanatory variables and production. 

However, DL models still encounter the following difficulties in production prediction:

(1) The inevitable occurrence of negative transfer in production prediction under insufficient data.

Although existing DL models have delivered satisfactory results, their performance heavily depends on having a sufficient quantity of samples. However, meeting this requirement may not be feasible in a newly developed shale gas block. Furthermore, the existing  models based on transfer learning (TL) ignore the problem of negative transfer.

(2) The inherent black-box nature of DL models.

The practical applicability of DL models is constrained by their inherent black-box nature, creating doubts among decision-makers about the validity of their results.

Establishing a reliable shale gas production prediction model under insufficient data, while avoiding negative transfer, is a valuable challenge. Therefore, we employ domain adaption TL and the addition of physical constrains to the DL models can address the above mentioned difficulties. TL is an effective technology to solve the problem of insufficient data. The concept of TL originates from the knowledge transfer mechanisms observed in human learning processes \cite{a35zhuang2020comprehensive35}. At its core, TL involves the application of knowledge and characteristics acquired from a source domain to improve the performance and efficiency of models in a target domain \cite{a36zhou2022time36}. To go into details, TL comprises two distinct stages: pre-training and fine-tuning \cite{a37ding2023machinery37}. During the pre-training stage, models are trained on source domain that are sufficiently endowed with labeled data and bear some relevance to the target domain. The primary objective of TL is to acquire feature representations with high generalization capabilities in this stage \cite{a38wang2021domain38}. On the other hand, the fine-tuning phase involves applying pre-trained models to the target domain and making necessary adjustments to align them with the demands of the new domain. Fine-tuning often includes self-adaptive updates to model parameters to ensure the preservation of universally applicable knowledge from the source domain while optimizing for task-specific information in the target domain \cite{a39yuan2023attention39}. The paramount objective of TL is to uncover commonalities and similarities between the source and target domains and then transfer knowledge from the former to the latter. The significance of this method lies in its ability to mitigate challenges stemming from data scarcity in the target domain, ultimately bolstering a model's generalization capacity. TL approaches have garnered substantial attention owing to their capacity to identify and leverage domain-agnostic features to ensure effective knowledge transfer. Especially in the face of insufficient data and distribution disparities, it is a potent tool for addressing real-world problems in the academic and industrial realms \cite{a40oyewole2022controllable40}.

TL has seen extensive application in various fields, including image classification \cite{a22hermessi2019deep22}, text classification \cite{a23griesshaber2020low23}, and computer vision \cite{a24fernando2013unsupervised24}. However, its application in production prediction, particularly for shale gas, remains limited. Niu et al. \cite{a25niu2023toward25} utilized transfer component analysis (TCA) and deep learning (DL) to predict static production and estimate maximum recovery in various shale gas wells. In a separate study, Niu et al. \cite{a56niu2023ensemble56} proposed an ensemble learning-based transfer learning method to enhance neural network performance for predicting shale gas production across different formations and blocks. Yet, these studies primarily focus on transferring knowledge regarding final static production. To the best of our knowledge, in terms of dynamic production prediction, the research is scarce. Traditional fine-tuning TL methods involve transferring knowledge between neural networks of identical structures. However, these methods often face the issue of negative transfer, especially when the target domain's data features significantly differ from those in the source domain. This challenge arises from traditional TL methods neglecting the differences in data distribution between the source and target domains. To address this, we propose a novel transfer learning methodology for shale gas production that leverages domain adaptation and physical constraints. This approach aims to mitigate negative transfer and enhance the interpretability of DL models.

\begin{itemize}
    \item  Leveraging the maximum entropy principle, a dynamic segmentation method is applied to divide a single source domain into multiple subsource domains, enhancing the diversity of the samples.
    \item Dynamically determining knowledge transfer is achieved through the Maximum Mean Discrepancy (MMD) and global average distance , effectively avoiding the problem of negative transfer from the perspective of data distribution.
    \item a hybrid model is proposed that utilizes MLP to incorporate drilling, completion,  and geological information as physical constraints,  combined with Transformer and attentional mechanism to improve the interpretability of the model.
 \end{itemize}

The remainder of this article is organized as follows: Section II introduces a detailed methodology of domain adaption in shale gas prediction production. Section III presents the proposed hybrid Transformer-MLP model based on physical constrains. Section IV conducts experimental validation and analyzes the results. Section V concludes the proposed methodology and outlines our future work. 

% methodology
\section{Methodology of domain adaption for shale gas production prediction}
\begin{figure}[ht]
    \centering
    \includegraphics[width=0.5\textwidth]{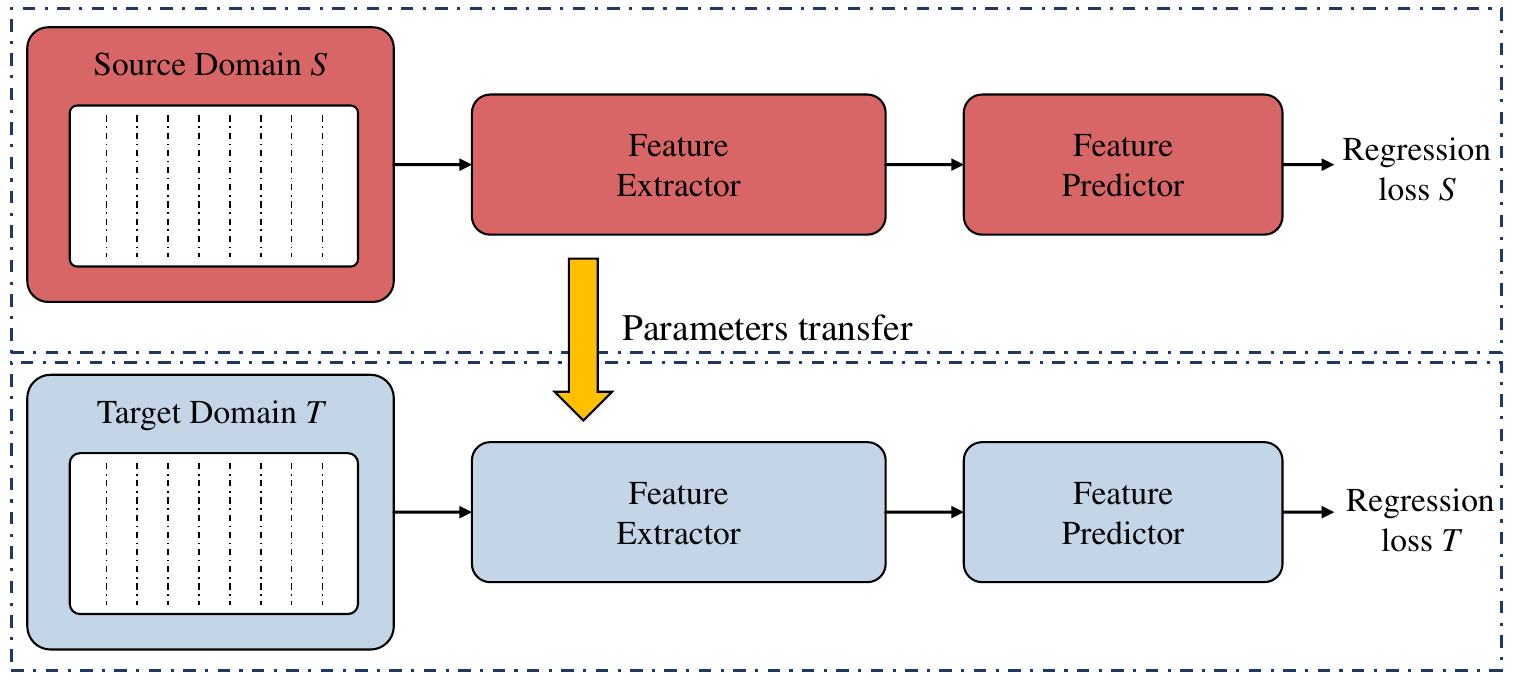}
    \caption{Traditional transfer learning model framework.}
    \label{fig:fig6}
\end{figure}

The traditional transfer learning framework is illustrated in \autoref{fig:fig6}. The target domain model undergoes migration by sharing parameters for feature extraction from the source domain model and subsequently retraining the feature extractor's parameters. However, this approach overlooks the distribution difference between the target and source domain data, inevitably resulting in negative transfer.

\begin{figure*}[htbp]
    \centering
    \includegraphics[width=1\textwidth]{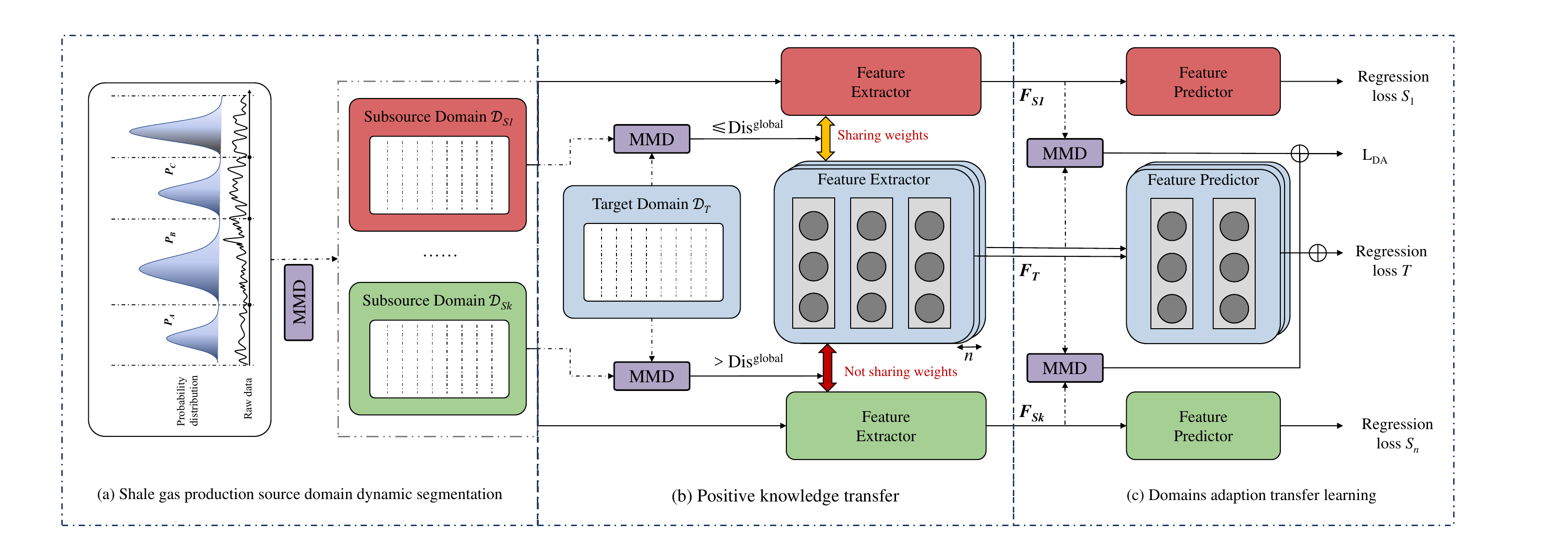}
    \caption{The proposed dynamic domain adaption framework.}
    \label{fig:fig2}
\end{figure*}

To address the problem, we propose a novel domain adaption TL methodology from the perspective of data distribution. The complete algorithm process is presented in \autoref{Algorithm1}. This method is primarily divided into three stages. Firstly, we utilize a dynamic segmentation algorithm to partition the accumulated historical production data from the shale gas source domain into multiple batches. Then, each of these batches, after splitting, is considered as an individual subsource domain. This is crucial as historical production data from different time periods may exhibit distinct distribution characteristics compared to the target domain production data. In the second stage, the model compares MMD distance and global average distance to identify subsource domain data that is similar to the target domain. It dynamically evaluates whether knowledge transfer is reasonable. In the final stage, we consolidate all transferable knowledge to maximize the utilization of information from the source domains, enabling domain adaptation. The proposed dynamic domain adaption framework is shown in \autoref{fig:fig2}.

\begin{algorithm}
  \caption{Domain adaptation shale gas production prediction}
  \label{Algorithm1}
  \KwIn{$D_S$ (source domain),$D_T$ (target domain)}
  \KwOut{the target domain shale gas production prediction model}
  \textbf{Initialization:} weights $w_{E_s}$, $w_{E_t}$ for source and target feature extractor $E_s$, $E_t$; parameters $\theta_{P_s}$, $\theta_{P_t}$ for source and target feature predictor $P_s$, $P_t$\;
  
  \textbf{Dynamic segmentation stage}\;
  Segment $D_S$ into $K$ subsource domains $\left\{\mathcal{D}_{S_i}\right\}_{i=1}^K$ through \autoref{equ:equ4}\;
  
  \For {$i=1$ to $K$}
  {calculate MMD distance $\operatorname{Dis}\left(\mathcal{D}_T, \mathcal{D}_{S_i}\right)$ as \autoref{equ:equ5}}

  Calculate the global average distance   through \autoref{equ:equ7}
  
  \textbf{Positive knowledge transfer stage}\;
  \For{$i=1$ to $K$}
  {
  Sample $x_s, y_s \sim D_{S_i}$ and $x_s, y_t \sim D_T$\;
  Feed samples to the feature extractor $E$\;
  \eIf{$\operatorname{Dis}\left(\mathcal{D}_T, \mathcal{D}_{S_i}\right) \leq \operatorname{Dis}^{\text {global }}$}
  { 
  sharing weights for $D_{S_i}$ and $D_T$,
  namely, $w_{E_t}=w_{E_s}$\;
  }
  {
  not sharing weights for $D_{S_i}$ and $D_T$, namely, $w_{E_t} \neq w_{E_s}$\;
  } 
  From feature extractor, get the time features \;
  ${\hat{x}}_s = E_s\left(x_s,w_{E_s}\right)$\;
  ${\hat{x}}_t = E_t\left(x_t, w_{E_t}\right)$\;

  \textbf{Domain adaptation transfer learning stage}\;
  Feed ${\hat{x}}_s$ and ${\hat{x}}_t$ to the feature predictor to get the subsource domain predicted values
  ${\hat{y}}_s = P_s\left({\hat{x}}_s;\theta_{P_s}\right)$ and target domain predicted values ${\hat{y}}_t = P_t\left({\hat{x}}_t;\theta_{P_t}\right)$\;
  
  Calculate the MMD distance between ${\hat{x}}_s$ and ${\hat{x}}_t$ as \autoref{equ:equ5}\;

    Calculate the domain adaptation loss $\mathcal{L}_{DA}(\theta)$ as \autoref{equ:equ9}\;
  
  Calculate the regression loss $\mathcal{L}_{regression}(\theta)$ for $P_s$ and $P_t$\;
    }

  Total loss $\mathcal{L}(\theta)={ \mathcal{L}_{regression}}(\theta)+ \mathcal{L}_{D A}(\theta)$\;
  Update the predicted values, the feature extractor $E$, and the feature predictor $P$ by minimizing the total loss\;
  \textbf{return} the target domain of the model
\end{algorithm}

\subsection{Shale gas production source domain data dynamic segmentation}\label{section:section2.1}
\begin{figure}[ht]
    \centering
    \includegraphics[width=0.5\textwidth]{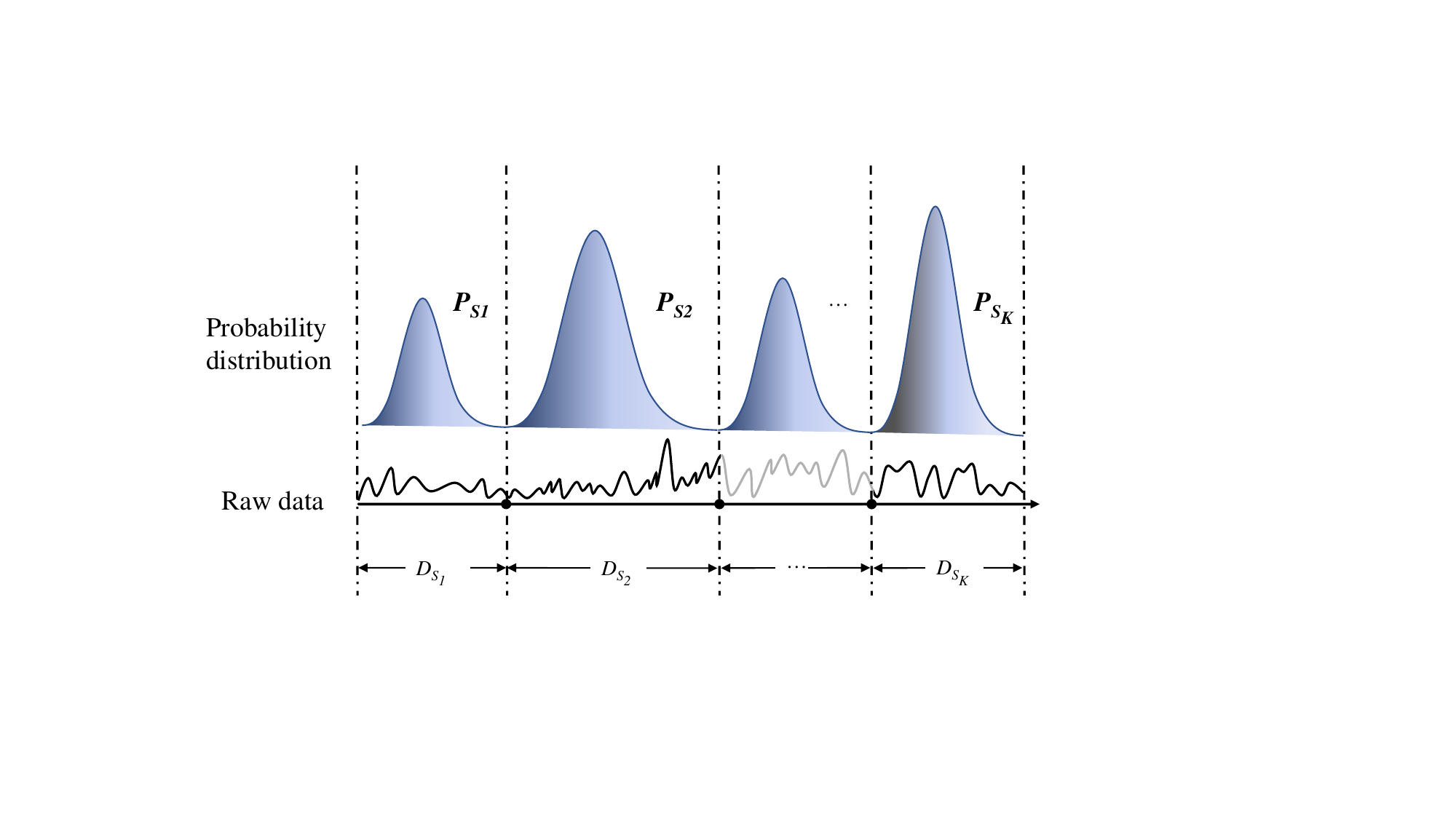}
    \caption{ Different data distribution in different time periods.}
    \label{fig:fig11}
\end{figure}

The production data from various historical periods exhibit distributional characteristics that different from those observed in the production data of the target domain. To extract useful information from the source domain data and enhance the diversity, our first objective is to dynamically segment the source domain, as shown in \autoref{fig:fig2}(a). The historical production data from the shale gas source domain is partitioned into $K$ batches. Each of these $K$ batches is considered as an individual subsource domain, allowing us to establish $K$ source domain models, as shown in \autoref{fig:fig11}. 

According to the principle of Maximum Entropy \cite{a41zhao2023maximum41}, leaveraging this approach proves advantageous in the realm of dynamic time series transfer. Maximizing shared knowledge within time series requires pinpointing the least similar time intervals to each other. Given the substantial diversity across these time periods, this method effectively bolsters the diversity of data \cite{a42du2021adarnn42}. To accomplish this objective, the dynamic segmentation is implemented by solving an optimization problem as follows:

\begin{equation}
\begin{aligned}
\label{equ:equ4}
& \max _{0<K \leq K_0} \max _{n_1, \cdots, n_K} \frac{1}{K} \sum_{1 \leq i \neq j \leq K} d\left(\mathcal{D}_i, \mathcal{D}_j\right), \\
& \text {s.t.} \forall i, \Delta_1<\left|\mathcal{D}_i\right|<\Delta_2 ; \sum_j\left|\mathcal{D}_i\right|=n.
\end{aligned}
\end{equation}
where $d$(.) is a distance metric. The parameters $\Delta_1$ and $\Delta_2$ are predefined to prevent trivial solutions. $K_0$ is a hyperparameter employed to mitigate over-splitting.

The above optimization problem \autoref{equ:equ4} aims to maximize the average distribution distance by searching for \textit{K} and their corresponding periods, with the goal of promoting diversity in the distribution within each period. To achieve this objective, we utilize MMD to calculate the average distribution distance. Briefly, MMD quantifies distribution differences by calculating the distance between the mean of the domain in the Reproducing Kernel Hilbert Space (RKHS). MMD is maximized when the mean difference reaches the largest value, indicating a greater similarity between the two domains when the MMD value is smaller. The MMD formula is shown below:

\begin{equation}
\label{equ:equ5}
\operatorname{MMD}\left(X_S, X_T\right)=\frac{1}{N_S} \sum_{N_S}^{i=1} \varphi\left(x_S^i\right)-\frac{1}{N_T} \sum_{N_T}^{j=1} \varphi\left(x_T^j\right)_H^2.
\end{equation}
where $\varphi$(.) is the Radial Basis Function \cite{a43buhmann2000radial43}, $N_S$ and $N_T$ are the number of samples for source and target domain.

Then, we apply a greedy algorithm \cite{a44vince2002framework44} to solve this optimization problem. After determining the split points, the source domain is divided into $K$ subsource domains. As a result, the resulting predictive model possesses enhanced generalization capabilities, promoteing diversity within each period.

\subsection{Positive knowledge transfer}

The secondary objective is to screen out the subsource domains similar to the target domain to ensure positive knowledge transfer, from the perspective of the data distribution, as shown in \autoref{fig:fig2}(b). Because not all the knowledge acquired from the subsource domains proves beneficial for learning in the target domain. We introduce the concept of global average distance to evaluate the similarity. The distances between each subsource domain $\mathcal{D}_{S_i}$ and the target domain $\mathcal{D}_{T}$ are as follows:

\vspace{-3pt}
\begin{equation}
\label{equ:equ6}
\operatorname{Dis}\left(\mathcal{D}_T, \mathcal{D}_{S_i}\right)=\operatorname{MMD}\left(\mathcal{D}_T, \mathcal{D}_{S_i}\right).
\end{equation}

Then, the method determines the transfer subsource domains by comparing the distance between $\operatorname{Dis}\left(\mathcal{D}_T, \mathcal{D}_{S_i}\right)$ and the global average distance.
When $\operatorname{Dis}\left(\mathcal{D}_T, \mathcal{D}_{S_i}\right) \leq \operatorname{Dis}^{\text {global }}$, the subsource domain data $\mathcal{D}_{S_i}$ is considered as transferable; otherwise, this $\mathcal{D}_{S_i}$  is automatically disregards. To balance the extreme value differences, we calculate the global average distance as follows:

\begin{flalign}
\label{equ:equ7}
	&\begin{array}{l}
		\operatorname{\text{Dis}}^{\text {global }} = \frac{{\sum\nolimits_1^{K - 2} {Dis{^i}} }}{{K - 2}},\left\{ {\max \left( {{{\text{Dis}}}\left( {{D_{\rm T}},{D_{{S_i}}}} \right)} \right)} \right\} \notin{\text{Dis}}{^i}. \\ 
	\end{array}&
\end{flalign}

\begin{figure*}[htbp]
    \centering
    \includegraphics[width=0.6\textwidth]{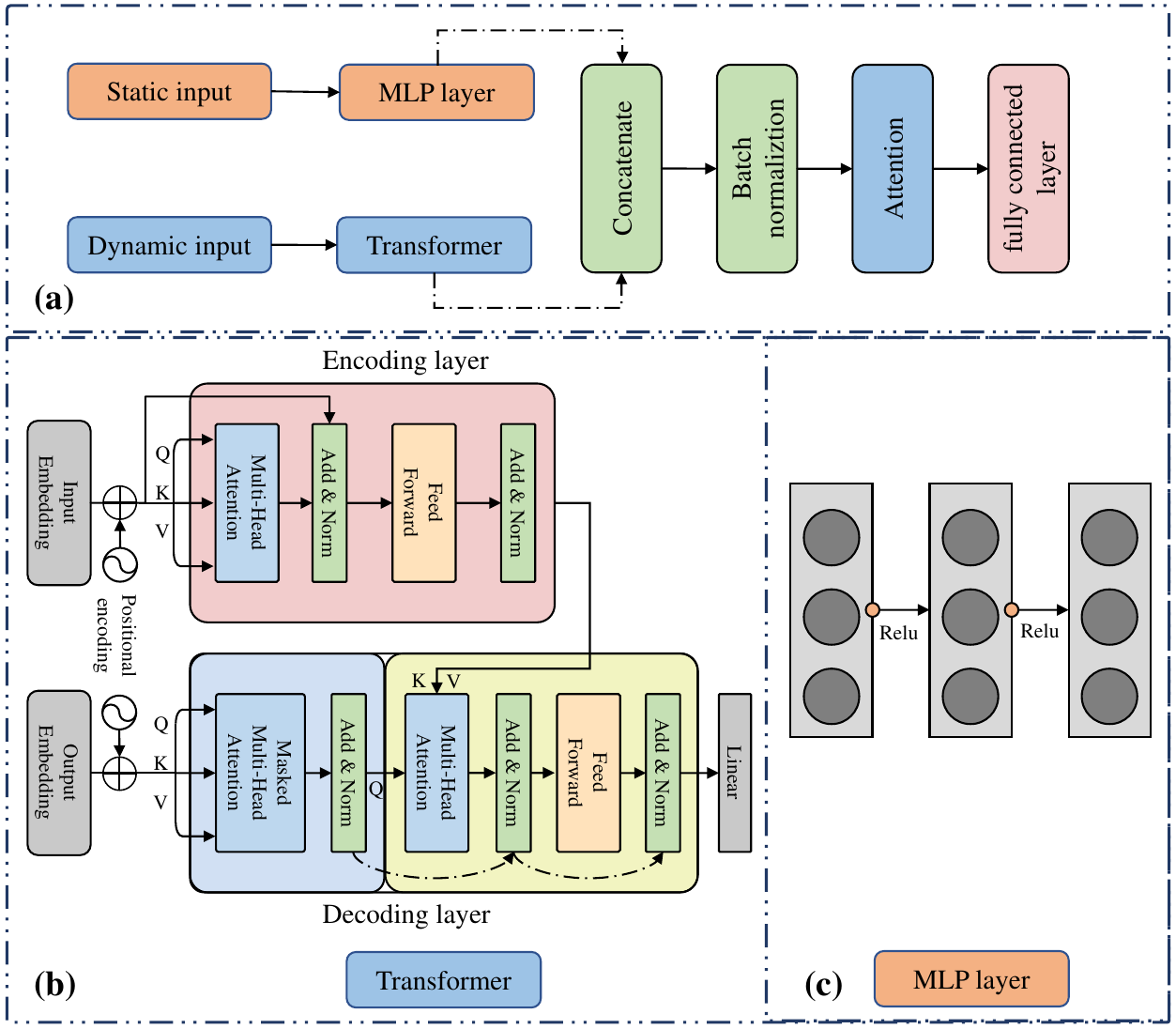}
    \caption{The Transformer-MLP model structure.}
    \label{fig:fig3}
\end{figure*}

\subsection{Domains adaption transfer learning}
The ultimate goal is to integrate all transferable knowledge for establishing a target domain model, as illustrated in \autoref{fig:fig2}(c). When considering domain adaptation feature mapping, we calculate the distance between the features mapping extracted from the subsource domains and the target domain. To enhance the prediction accuracy of the target domain, the target feature distribution should closely align with the feature distribution of the source domains. We then minimize the prediction loss of each source domain and the target model. The loss function of dynamic TL process between different structures can be described as follows:

\vspace{-3pt}
\begin{equation}
\label{equ:equ8}
\mathcal{L}(\theta)={ \mathcal{L}_{regression}}(\theta)+ \mathcal{L}_{D A}(\theta).
\end{equation}

The domain adaption loss from the feature mapping is as follows:

\vspace{-3pt}
\begin{equation}
\label{equ:equ9}
	\begin{array}{c}
		\mathcal{L}_{D A}(\theta) = \operatorname{MMD}\left( {{F_{S1}},{F_T}} \right) +\operatorname{MMD}\left( {{F_{S2}},{F_T}} \right) +,  \\ 
		\cdots  + \operatorname{MMD}\left( {{F_{SK}},{F_T}} \right). \\ 
	\end{array}
\end{equation}
where $F_{S i}$ are the features extracted from the source domains, $F_T$ represent the features extracted from the target domain.

The regression loss of each source domain and the target model is follows:

\begin{equation}
\label{equ:equ10}
\begin{aligned}
L_{\text {regression }} & =\frac{1}{\left|D_T\right|} \sum_{i=1}^{\left|D_T\right|} {\text {MSE}}\left(y_{T i}, M\left(x_{T i} ; \theta\right)\right), \\& + \frac{1}{K} \sum_{j=1}^K \frac{1}{\left|D_{S j}\right|} \sum_{i=1}^{\left|D_{S i}\right|} {\text {MSE}}\left(y_{S i}^j, M\left(x_{S i}^j ; \theta\right)\right).
\end{aligned}
\end{equation}
where MSE is the Mean Squared Error \cite{a45das2004mean45}, $M$ denotes the proposed hybrid model, $y$ represents the true value, $x$ are the inputs, and $\theta$ refers to the model parameters.

\section{ The hybrid Transformer-MLP prediction model  based on physical constraints}
The inherent black-box nature of DL models raises doubts among decision-makers about the validity of their results, limiting their practical applicability. To address the problem, we propose a hybrid Transformer-MLP model, serving as feature extractor. This model incorporates drilling, completion, and geological information as physical constrains to ensure model interpretability. 
\subsection{Transformer}
The Transformer model was first proposed by Vaswani et al. for natural language processing \cite{a29vaswani2017attention29}. It relies primarily on the self-attention mechanism to enhance its ability to capture primary dependencies between input and output \cite{a30wang2023convolutional30}. The Transformer architecture consists of two core components: the Encoder and Decoder, as shown in \autoref{fig:fig3}(b).

The Encoder consists of stacked units designed to capture dependencies within the input sequence, incorporating multihead attention and a feedforward neural network. The output of one unit feeds into the next, forming a layered propagation of information.  To address gradient problems, residual connections integrate the original input with the unit output. Layer normalization is employed to ensures stable training \cite{a31han2021transformer31}. The Decoder, responsible for generating final predictions, is composed of multiple units, each featuring two multi-head attention layers and a feedforward neural network. Specialized self-attention within Decoder units captures the Encoder-Decoder relationship. Residual connections and layer normalization maintain information flow and stability \cite{a32wen2022transformers32}.

Attention is a crucial component in the Transformer, enabling features to interact with each another and generate more informative and expressive representations \cite{a33huang2019dsanet33}. The attention mechanism takes three essential elements: the query matrix $\mathbf{Q}$, the key matrix $\mathbf{K}$, and the value matrix $\mathbf{V}$. The query matrix $\mathbf{Q}$ is used to compare with the key matrix $\mathbf{K}$, resulting in a set of weights that mirror the relationships between various queries and keys. A higher weight indicates a stronger correlation. These weights can then be used to modulate the value matrix $\mathbf{V}$, generating a weighted average for each query. Consequently, the attention score is calculated by multiplying each weight by the corresponding value vector and subsequently adding these weighted value vectors. The scaled dot-product attention is as follows:

\vspace{-3pt}
\begin{equation}
\label{equ:equ1}
\mathbf{Attention}(\mathbf{Q}, \mathbf{K}, \mathbf{V})={\text{ softmax }}\left( \frac{\mathbf{Q K}^{\top}}{\sqrt{d_k}}\right) \mathbf{V}.
\end{equation}
where $\mathbf{Q}$, $\mathbf{K}$, and $\mathbf{V}$ are generally obtained by applying some transformations to the original inputs; $d_k$ represents the dimensions of keys.

The Transformer usually adopts multihead attention \cite{a34gao2022tgdlf234}. The multihead attention mechanism differs from a single self-attention mechanism in that it compares each query to a set of key vectors from multiple representation subspaces. Performs this calculation multiple times, not just once. The output matrix of the $i$ th self-attention mechanism is defined as $\mathbf{H e a d}_i (i=1,  2, \ldots, n)$, which is computed as:

\vspace{-3pt}
\begin{equation}
\label{equ:equ2}
\mathbf{H e a d}_i=\mathbf{Attention}\left(\mathbf{Q}^{\prime} \mathbf{W}_i^{\mathrm{Q}}, \mathbf{K}^{\prime} \mathbf{W}_i^{\mathrm{K}}, \mathbf{V}^{\prime} \mathbf{W}_i^{\mathrm{V}}\right) .
\end{equation}

\vspace{-5pt}
\begin{equation}
\label{equ:equ3}
\mathbf{multi}-\mathbf{head}={\text{ Concat }}\left(\mathbf{H e a d}_1, \ldots, \mathbf{H e a d}_n\right) \mathbf{W}^o.
\end{equation}
where $\mathbf{multi}-\mathbf{head}$ is the output matrix of multihead attention, ${\text{ Concat }}$(.) is the concatenation operation, $n$ is the number of heads, and $\mathbf{W}^o$ is the parameter matrix.
\subsection{Hybird Transformer-MLP deep learning model}
By incorporating physical constraints, we propose a novel hybrid model called the Transformer-MLP model, depicted in \autoref{fig:fig3}(a). This model combines the capabilities of a Transformer for extracting temporal features and an MLP for incorporating static physical constraints. It utilizes an attention mechanism to emphasize crucial features. Different from the conventional DL production prediction models, this hybrid model processes static physical information and dynamic production data simultaneously. This approach enables us to extract physically meaningful features from the input data, resulting in greater reliability and interpretability compared to traditional models that solely rely on time-based structures.

The dynamic feature structure of the proposed hybrid model is illustrated in \autoref{fig:fig3}(b). Initially, gas and water production data from multiple time series are converted into matrices using a time window and then fed into the Transformer. Subsequently, static physical information, including drilling, completion, and geological features, is input into the MLP. The number of neurons is determined by the number of static input features in the static input layer. The dynamic data within the sliding window and static data are defined as the following equations:

\begin{figure*}[htbp]
    \centering
    \includegraphics[width=0.8\textwidth]{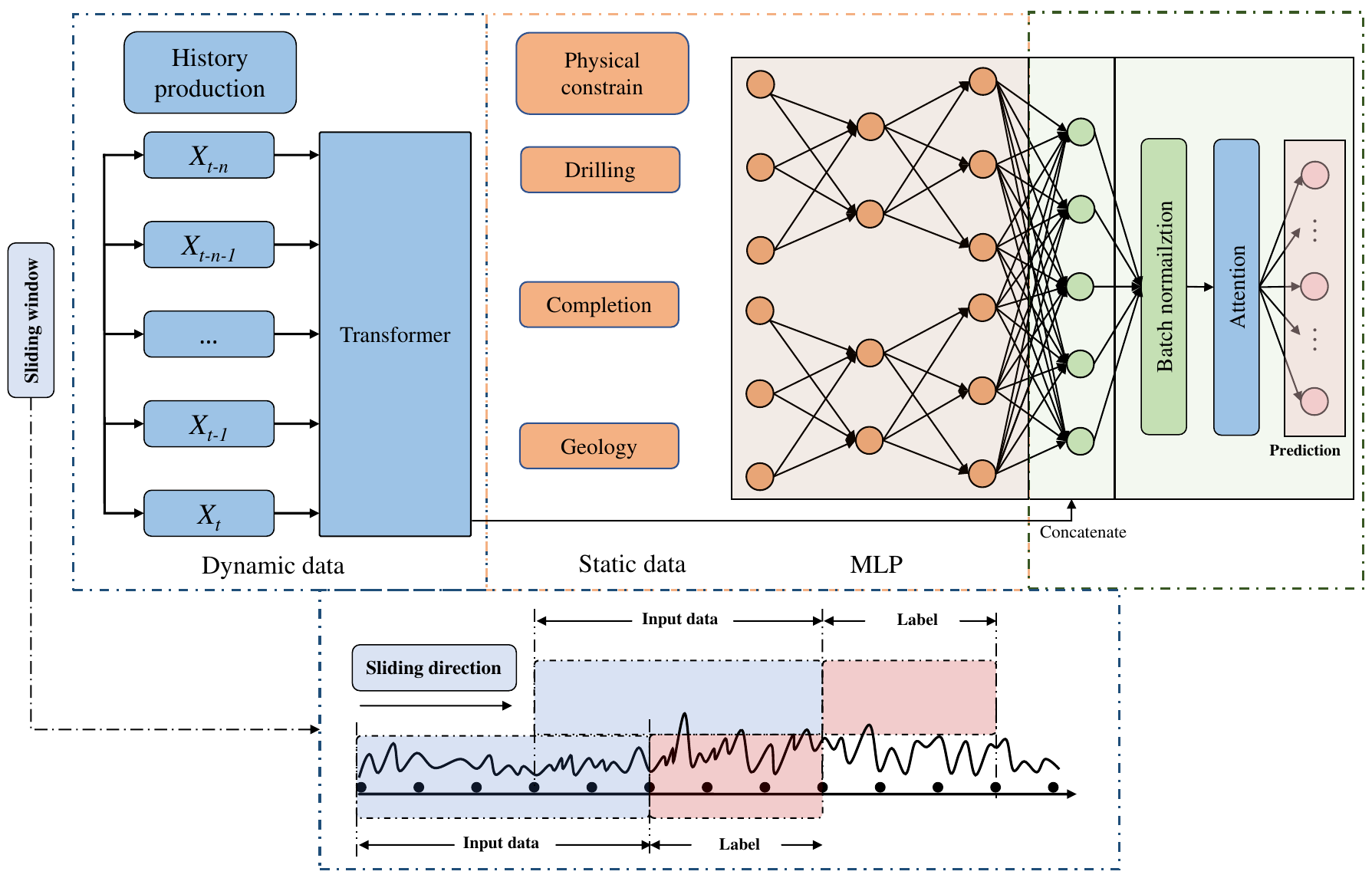}
    \caption{Inputs and outputs diagram of the proposed model.}
    \label{fig:fig12}
\end{figure*}

\begin{equation}
\label{equ:equ10}
X_d=\left[\begin{array}{cccc}
x_{d, 1}^1 & x_{d, 2}^1 & \cdots & x_{d, t}^1 \\
x_{d, 1}^2 & x_{d, 2}^2 & \cdots & x_{d, t}^2
.\end{array}\right]
\end{equation}

\vspace{-5pt}
\begin{equation}
\label{equ:equ11}
X_s=\left(\begin{array}{llll}
x_{\mathrm{s}}^1 & x_{\mathrm{s}}^2 & \cdots & x_{\mathrm{s}}^m
\end{array}\right)^{\top}.
\end{equation}

As shown in \autoref{fig:fig12}, the process begins with dimensionality expansion of static data through a linear layer. This step is crucial in preventing the feature dimension of the time series from becoming excessively large, thereby avoiding the risk of the model neglecting features from the static input component. Following this, the MLP architecture comprises three linear layers, each separated by Rectified Linear Unit (ReLU) activation functions. Previous research \cite{a25niu2023toward25} suggest that a three-layer structure (including two hidden layers) is sufficient for approximating complex function structures. While deeper network architectures can enhance learning efficiency and model performance to some extent, they may also increase the risk of overfitting.

After combining the outputs from both the MLP and Transformer, we apply batch normalization before inputting them into the attention layer. The attention layer accesses the significance of crucial dynamic and static features. Ultimately, a linear layer is employed to predict gas and water production for the target time period. This structured approach enhances the logical coherence of our model, making it more competitive, especially in shale gas production prediction scenarios under insufficient data.

\begin{figure*}[htbp]
    \centering
    \includegraphics[width=0.5\textwidth]{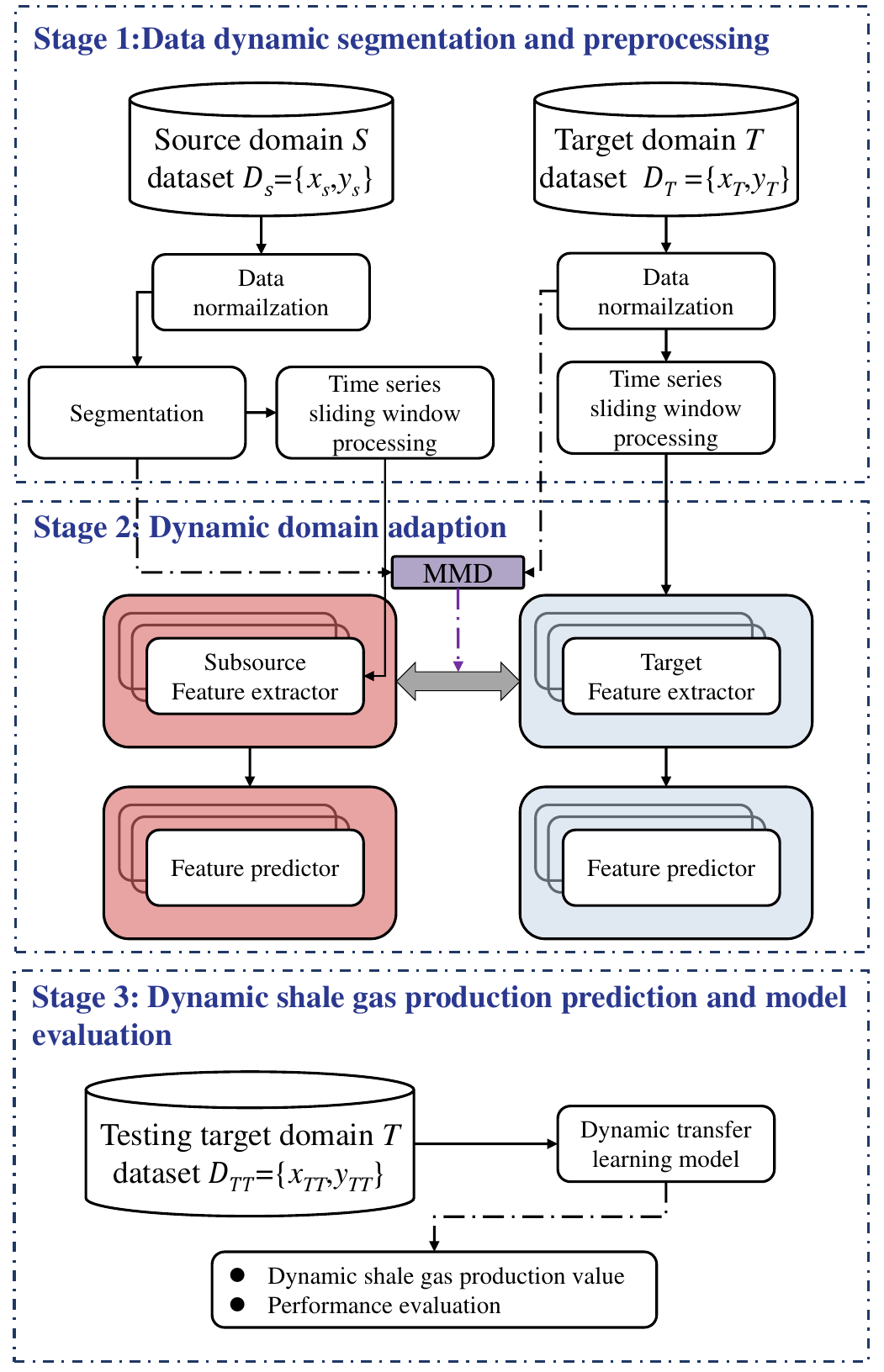}
    \caption{The flowchart of the proposed method.}
    \label{fig:fig4}
\end{figure*}

The workflow of the proposed methodology can be divided into the following three steps, as illustrated in \autoref{fig:fig4}.

Step 1: Data preprocessing and dynamic segmentation. Begin by collecting shale gas source and target domain data. Then, normalize the data using the min-max scaling normalization \cite{a46patro2015normalization46} to scale all values within the range (0, 1). Next, apply the dynamic segmentation algorithm described in \autoref{section:section2.1} to partition the source domain into $K$ subdomains. Finally, preprocess the production data using the sliding window method.

Step 2: Domain adaptation transfer learning. Utilizing the proposed Transformer-MLP model, we extract the intricate relationship between input and production. Subsequently, we determine the transfer by assessing the similarity between each subsource domain and the target domain, along with the global average distance. We consolidate transferable knowledge to establish a model for the target domain. Lastly, we employ grid search to optimize the model hyperparameters.

Step 3: Shale gas and water production prediction and model evaluation. We input the test set from the target domain into the trained model using the optimal model configuration to obtain predictions. Since the input data has been normalized, we inversely normalize the predictions to generate final forecasts that correspond to the original observations. Finally, we evaluate the model based on predefined evaluation criteria.

\section{Experiments}
To validate the effectiveness and superior capabilities of the proposed methodology in shale production prediction, we conducted extensive experiments using datasets from the southwestern region of China. The experiments are divided into two groups. In the first set of experiments, we compared the performance of the hybrid model with DL models that have previously demonstrated excellent performance in production prediction. In the second set of experiments, we conducted a comparative study contrasting dynamic domain adaptation with traditional TL to demonstrate the accuracy and robustness of our methodology.

\subsection{Dataset description and experiment setting}
We utilize production datasets from two actual shale gas production blocks, Block A and Block B, located in the southwest region of China, to assess performance. Block A's primary production zone spans depths from 2500 to 4000 meters, encompassing a diverse range of reservoir types. This area is characterized by well-developed natural fractures, showcasing diversity in fracture orientations, indicative of a mature shale gas production region. On the other hand, Block B boasts a formation pressure coefficient exceeding 1.8, along with elevated gas content and gas saturation, signifying favorable conditions for shale gas preservation. The organic content and porosity of the shale in Block B are generally similar to those in Block A, albeit with slightly higher gas saturation and gas content. Block B, as one of the most recently developed shale gas blocks, has emerged as a key player in future shale gas production.

Data collection included 268 shale gas wells in Block A and 52 shale gas wells in Block B. To ensure data continuity and model stability, we selected shale gas wells with longer production histories, designating Block A as the source domain and Block B as the target domain. For each well, a time window was applied to process production data from the last 120 days, including wellhead pressure (MPa), flowline pressure (MPa), gas production (10$^4$ m$^3$/d), and water production (m$^3$/d) as inputs. Following the time window algorithm, Block A yielded approximately 200,000 data samples, while Block B resulted in around 25,000. Additionally, we incorporated static drilling factors (total length and well volume), completion parameters (fracturing segment length, number of stages, fracturing fluid intensity, intensity of sand, and displacement), and geological factors (TOC, horizontal minimum principal stress, and average porosity), as depicted in \autoref{tab:tab1}. These static factors, highlighted in the literature \cite{a25niu2023toward25,a26lopez2023development26, a56niu2023ensemble56, a57syed2022smart57, a58hui2021machinea58}, are recognized as key contributors influencing production. Ultimately, gas and water production data for the subsequent 30 days served as the outputs

% \usepackage{multirow}

% \usepackage{multirow}
\begin{table*}[width=1\textwidth,htbp]
\caption{Static data statistics of data set}
\label{tab:tab1}
\resizebox{\textwidth}{!}{
\begin{tabular}{llllllll}
\hline
\multirow{2}{*}{Category} & \multirow{2}{*}{Symbol} & \multirow{2}{*}{Parameter}          & \multirow{2}{*}{Unit} & \multicolumn{2}{l}{Block A} & \multicolumn{2}{l}{Block B} \\ \cline{5-8} 
                          &                         &                                     &                       & Mini            & Max             & Mini            & Max             \\ \hline
Drilling                  & Lw                      & Total length                        & m                     & 2240            & 6165            & 3826            & 6460            \\ \hline
                          & Wv                      & Wellbore volume                     & m                     & 20.6            & 64.3            & 40.1            & 66.2            \\ \hline
Completion                & Fsl                     & Fracturing Segment Length           & m                     & 511             & 2576            & 1058            & 2301            \\ \hline
                          & Ns                      & Number of stages                    & \#                    & 4               & 36              & 25              & 34              \\ \hline
                          & Fsi                     & Fracturing Fluids Intensity         & m$^3$/m                 & 16.5            & 28.6            & 21.6            & 36.76           \\ \hline
                          & Is                      & Intensity of Sand                   & t/m                   & 0.54            & 2.98            & 1.69            & 3.68            \\ \hline
                          & Ad                      & Displacement                        & m$^3$/min                & 10.2            & 16.8            & 11.0            & 18.2            \\ \hline
Geology                   & TOC                     & Total organic carbon                & \%                    & 3.2             & 7.1             & 2.4             & 5.6             \\ \hline
                          & $\sigma_h$              & Horizontal minimum principal stress & MPa                   & 65.4            & 99.6            & 78.3            & 100.8           \\ \hline 
                          & $\phi$                  & Average porosity                    & \%	           	    & 5.1	          &8.9	            &2.5	          &5.3              \\ \hline
\end{tabular}}
\end{table*}

To evaluate the model's performance, we employ three metrics: the root mean squared error (RMSE), the mean absolute error (MAE), and the R-squared (R$^2$) \cite{a45das2004mean45}. These metrics are calculated as follows:

\begin{equation}
\mathrm{RMSE}=\sqrt{\frac{1}{N} \sum_{i=1}^N\left(y_i-\hat{y}_i\right)^2}
\end{equation}

\begin{equation}
\mathrm{MAE}=\frac{1}{N} \sum_{i=1}^N\left|y_i-\hat{y}_i\right|
\end{equation}

\begin{equation}
\mathrm{R}^2=1-\frac{\sum_{i=1}^N\left(y_i-\hat{y}_i\right)^2}{\sum_{i=1}^N\left(y_i-\bar{y}\right)^2}.
\end{equation}
where \textit{N} is the number of test samples, and $y_i$ , $\hat{y}_i$   are the $i$th real producation data and the corresponding forecast respectively.

\subsection{Comparison of various DL methods}\label{section:section4.2}

In this subsection, we compare the performance of the Transformer-MLP with previously excellent DL models. Additionally, we introduce a physical constraints component into the baseline models to assess the contribution of physical constraints. To validate the performance of the proposed hybrid model, our study uses Block A as the dataset since the dataset of Block A is more comprehensive and representative compared to Block B. The input dimension of dynamic data is 4, the input dimension of static data is 10, and the output dimension is 2, representing daily gas and water production, respectively. We partition the data into three sets: 70\% for training, 20\% for validation, and the remaining 10\% for testing

\begin{table*}[width=0.75\textwidth, htbp]
    \caption{the range of hyperparameter search}
    \label{tab:tab2} 
    \resizebox{0.75\textwidth}{!}{
    \begin{tabular}{lll}
    \hline
    Model name                       & hyperparameter                    & search                 \\ \hline
    \multirow{2}{*}{Bi-LSTM\cite{a28du2023enhanced28}}         & LSTM units                        & \{\textbf{50},100,150,200,300\} \\ \cline{2-3} 
                                     & Number of layers                  & \{2,\textbf{4},6,8,10\}      \\ \hline
    \multirow{3}{*}{Bi-LSTM-MLP}     & LSTM units                        & \{\textbf{50},100,150,200,300\} \\ \cline{2-3} 
                                     & Number of layers                  & \{\textbf{2},4,6,8,10\}      \\ \cline{2-3} 
                                     & The number of neurons in MLP      & \{20,\textbf{40},60,60,100\}    \\ \hline
    \multirow{2}{*}{LSTM\cite{a51huang2022well51}}            & LSTM units                        & \{50,\textbf{100},150,200,300\} \\ \cline{2-3} 
                                     & Number of layers                  & \{2,4,\textbf{6},8,10\}      \\ \hline
    \multirow{3}{*}{LSTM-MLP}        & LSTM units                        & \{\textbf{50},100,150,200,300\} \\ \cline{2-3} 
                                     & Number of layers                  & \{2,4,\textbf{6},8,10\}      \\ \cline{2-3} 
                                     & The number of neurons in MLP      & \{20,40,\textbf{60},80,100\}    \\ \hline
    \multirow{2}{*}{TCN\cite{a20li2022predicting20}}                     
                                     & The number of convolution kernels & \{$2^2$,$2^3$,$\bm{2^4}$,$2^5$,$2^6$\}     \\ \cline{2-3} 
                                     & The kernel size                   & \{\textbf{2},3,5,7,9\}            \\ \hline
    \multirow{3}{*}{TCN-MLP}         & The number of convolution kernels & \{$2^2$,$\bm{2^3}$,$2^4$,$2^5$,$2^6$\}     \\ \cline{2-3} 
                                     & The kernel size                   & \{\textbf{2},3,5,7,9\}            \\ \cline{2-3} 
                                     & The number of neurons in MLP      & \{20,40,\textbf{60},60,100\}    \\ \hline
    \multirow{3}{*}{Transformer\cite{a21zhang2023evaluating21}}     & The number of heads               & \{2,4,6,\textbf{8},10\}         \\ \cline{2-3} 
                                     & The number of sub-Encoder layers  & \{2,4,8,\textbf{12},16\}        \\ \cline{2-3} 
                                     & The number of sub-Decoder layers  & \{2,4,\textbf{8},12,16\}        \\ \hline
    \multirow{4}{*}{Transformer-MLP} & The number of heads               & \{2,4,\textbf{6},8,10\}         \\ \cline{2-3} 
                                     & The number of sub-Encoder layers  & \{2,\textbf{4},8,12,16\}        \\ \cline{2-3} 
                                     & The number of sub-Decoder layers  & \{2,\textbf{4},8,12,16\}        \\ \cline{2-3} 
                                     & The number of neurons in MLP      & \{\textbf{20},40,60,60,100\}    \\ \hline
    \end{tabular}}
\end{table*}
 
\begin{table*}[width=0.8\textwidth, htbp]
    \caption{Comparison of prediction performance of all models on various DL methods}
    \label{tab:tab3}
    \resizebox{0.8\textwidth}{!}{
    \begin{tabular}{lllllll}
    \hline
    Methods      & Gas    &           &           & Water  &          &         \\ \hline
                 & R$^2$     & RMSE(m$^3$)  & MAE(m$^3$)   & R$^2$     & RMSE(m$^3$) & MAE(m$^3$) \\ \hline
    Bi-LSTM \cite{a28du2023enhanced28} & 0.8221  & 3115.1186 & 1547.1181 & 0.7898 & 43.6590   & 36.3918 \\ \hline
    Bi-LSTM-MLP  & 0.8767 & 2572.0346 & 1460.4256 & 0.8751 & 36.4708  & 29.9762 \\ \hline
    LSTM \cite{a51huang2022well51}    & 0.7682 & 4656.8327 & 3905.8025 & 0.7727 & 69.8927  & 60.5043 \\ \hline
    LSTM-MLP     & 0.8077 & 3995.2728 & 3310.4915 & 0.7932 & 61.6440   & 52.304  \\ \hline
    TCN \cite{a20li2022predicting20}     & 0.8414 & 3408.5745 & 2970.0212 & 0.8114 & 49.0224  & 41.0815 \\ \hline
    TCN-MLP      & 0.8609 & 2518.2191 & 2225.2538 & 0.8687 & 40.7030   & 34.1386 \\ \hline
    Transformer \cite{a29vaswani2017attention29}  & 0.8595 & 2098.2323 & 1280.5308 & 0.8278 & 31.9330   & 26.2273 \\ \hline
    \textbf{Our model}    & 0.9257 & 1655.0766 & 903.9395  & 0.9081 & 25.1806  & 20.2181 \\ \hline
    \end{tabular}}
\end{table*}

Hyperparameters are crucial to ensure a fair comparison between different models. All models are trained using similar training hyperparameters. For example, we apply the same fundamental training epochs, with a training duration of 1000 epochs, to train all models. Furthermore, we use a grid search approach to determine the batch size for all models, with a search range of \{32, 64, 128, 256\}. The learning rate search range includes \{1e-3, 1e-4, 1e-5, 1e-6\}, and the optimizer search range includes \{Adam \cite{a47kingma2014adam47}, SGD \cite{a48landro2020mixing48}, RMSprop \cite{a49kumar2020malaria49}, LBFGS \cite{a50alpak2023machine50}\}. \autoref{tab:tab2} displays the range of different model hyperparameter searches and the best results.

\begin{figure*}[htbp]
    \centering
    \includegraphics[width=0.85\textwidth]{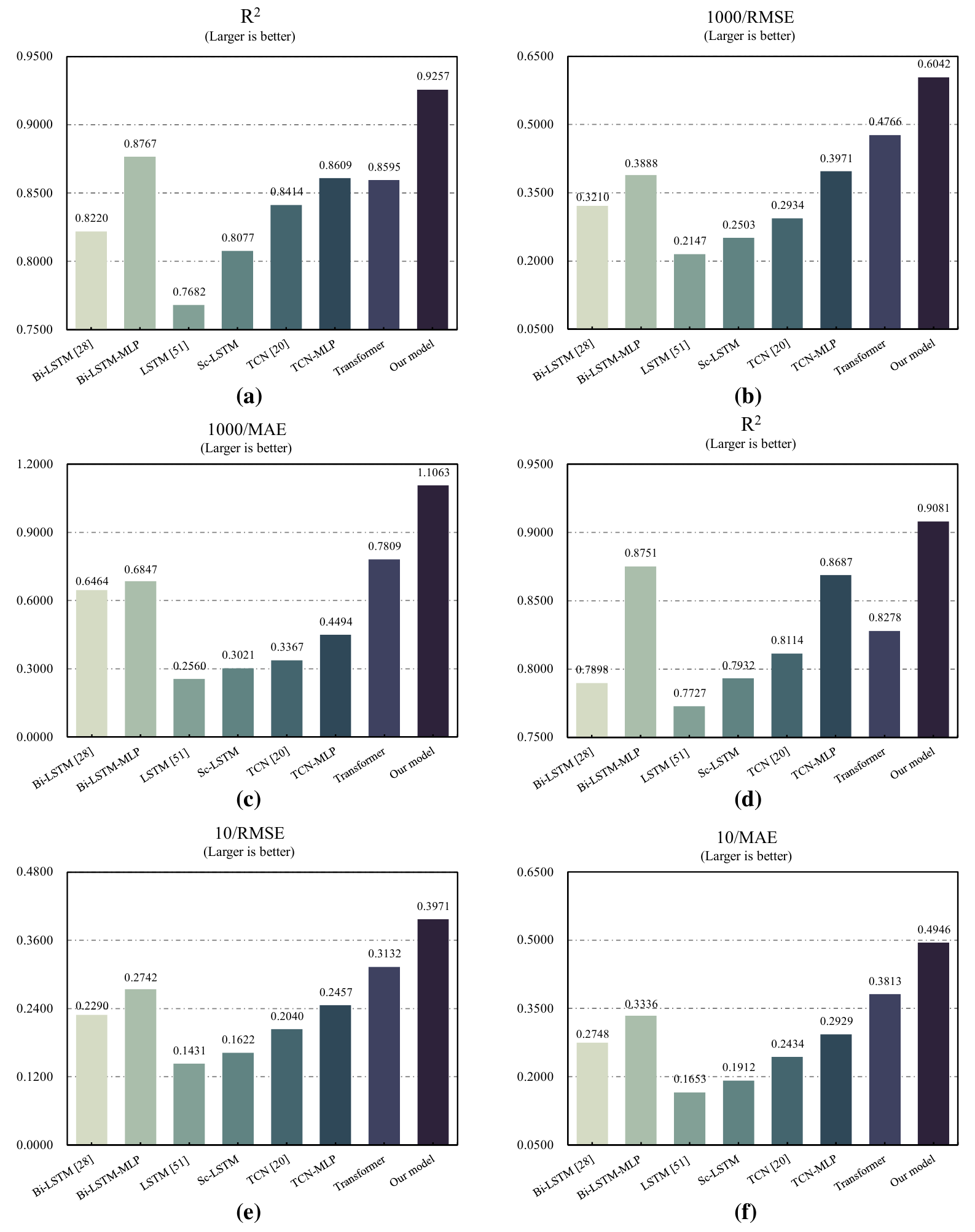}
    \caption{Comparison of the prediction performance for different models. Fig. 7 (a, b, c) denotes the R$^2$ scores, 1000 / RMSE, 1000 / MAE for the prediction of gas production in the set of source domain tests, respectively. Fig. 7 (d, e, f) denote R$^2$ scores, 10/RMSE, 10/MAE for water production prediction on the source domain test set, respectively.}
    \label{fig:fig5}
\end{figure*}

After conducting experiments with the source domain dataset, the results for all models, including RMSE, MAE, and R$^2$ values, are presented in \autoref{tab:tab3}. These results are further visualized in \autoref{fig:fig5}. It is evident  that the proposed model outperforms others in predicting gas and water production. Below are detailed analysis results:

(1) The comparison of prediction results clearly demonstrates that the inclusion of physical constraints enhances the predictive performance of the model. Taking RMSE as an example, for models augmented with physical information, the average values for gas and water production prediction scenarios are 2685.1m$^3$/d and 40.1m$^3$/d, respectively. These values are significantly lower than those of DL models without physical information, which have average RMSE values of 3319.6m$^3$/d and 48.6m$^3$/d for gas and water production prediction scenarios. This indicates that physical constraints effectively assist the model in comprehending dynamic production changes and future development trends.

(2) When comparing the outcomes of fusing physical information with different DL models, our proposed Transformer-MLP stands out with the lowest RMSE and MAE values, as well as the highest R$^2$ values, signifying its superior predictive accuracy. For instance, the average R$^2$ value of the proposed model in gas and water production prediction scenarios is 0.9169, markedly surpassing the average R$^2$ values of other DL models in the same scenarios, such as LSTM-MLP (0.8005), TCN-MLP (0.8648), and Bi-LSTM-MLP (0.8759). These results highlight that, in comparison to various DL-based models, our proposed Transformer-MLP is the optimal choice for integrating physical information and dynamic data in production prediction. The model's attention mechanism enables it to handle relevant information in the input data more effectively, thereby enhancing model performance and accuracy. It aids the model in focusing on critical input components while reducing the impact of irrelevant information.

\subsection{Performance of domain adaption for shale gas production prediction}

To evaluate the effectiveness of the proposed methodology, we employ data from both the source domain (Block A) and the target domain (Block B). We compare the results with those obtained using traditional fine-tuning transfer learning methods. Additionally, we utilize the Transformer-MLP as the feature extractor and the fully connected layer as the feature predictor.  It's important to highlight that, to maintain consistency and avoid potential bias in the research outcomes, all hyperparameters are selected based on the results presented in \autoref{section:section4.2}.

\begin{figure*}[htbp]
    \centering
    \includegraphics[width=1\textwidth]{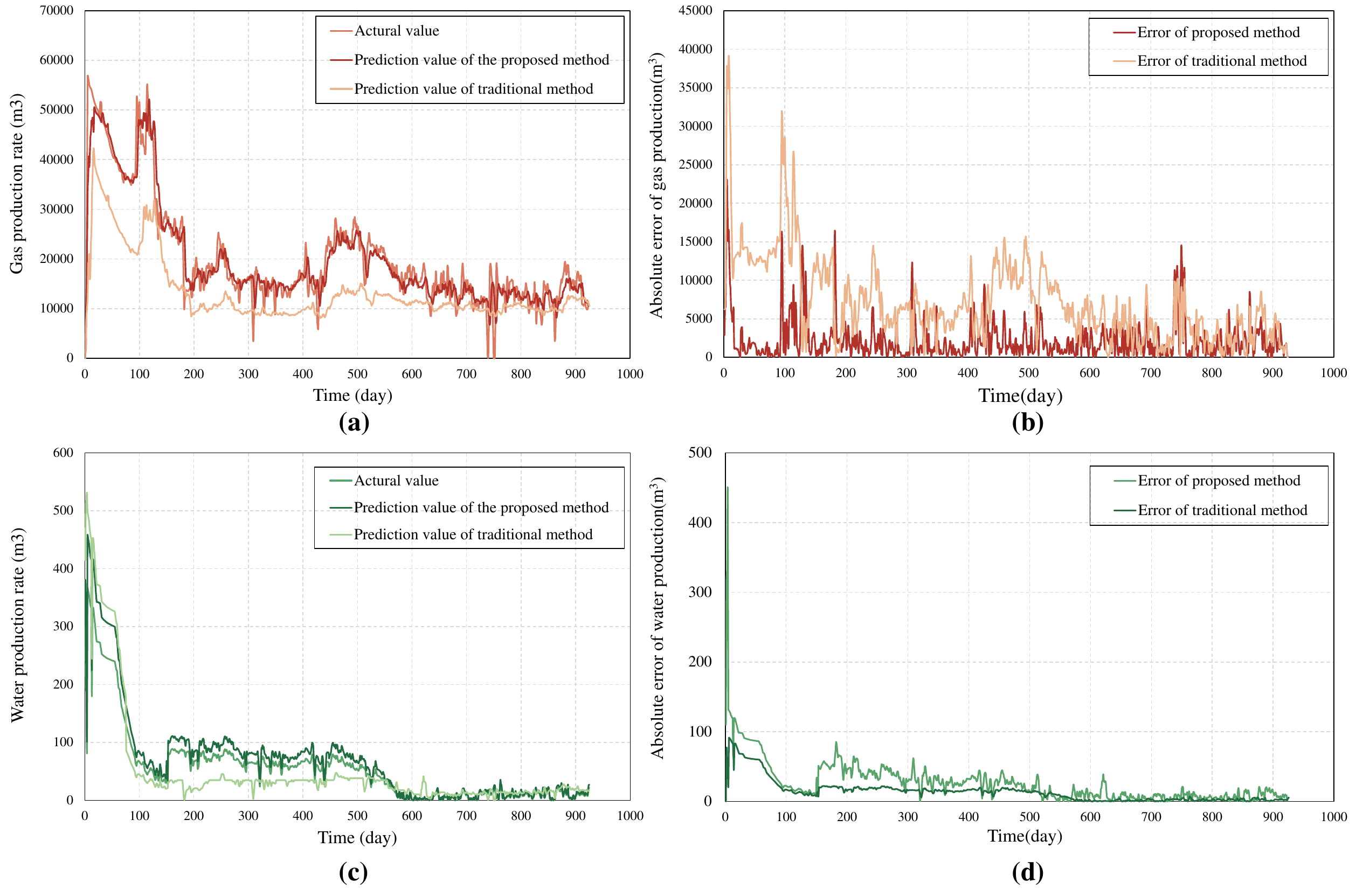}
    \caption{Comparison of the gas and water production prediction results between the proposed method and traditional method for a typical well in target domain. Fig. 8 (a,c) denotes the comparison of the gas and water production result. Fig. 8 (b,d) illustrates the comparison of the gas and water production errors.}
    \label{fig:fig7}
\end{figure*}

\begin{figure*}[htbp]
    \centering
    \includegraphics[width=0.8\textwidth]{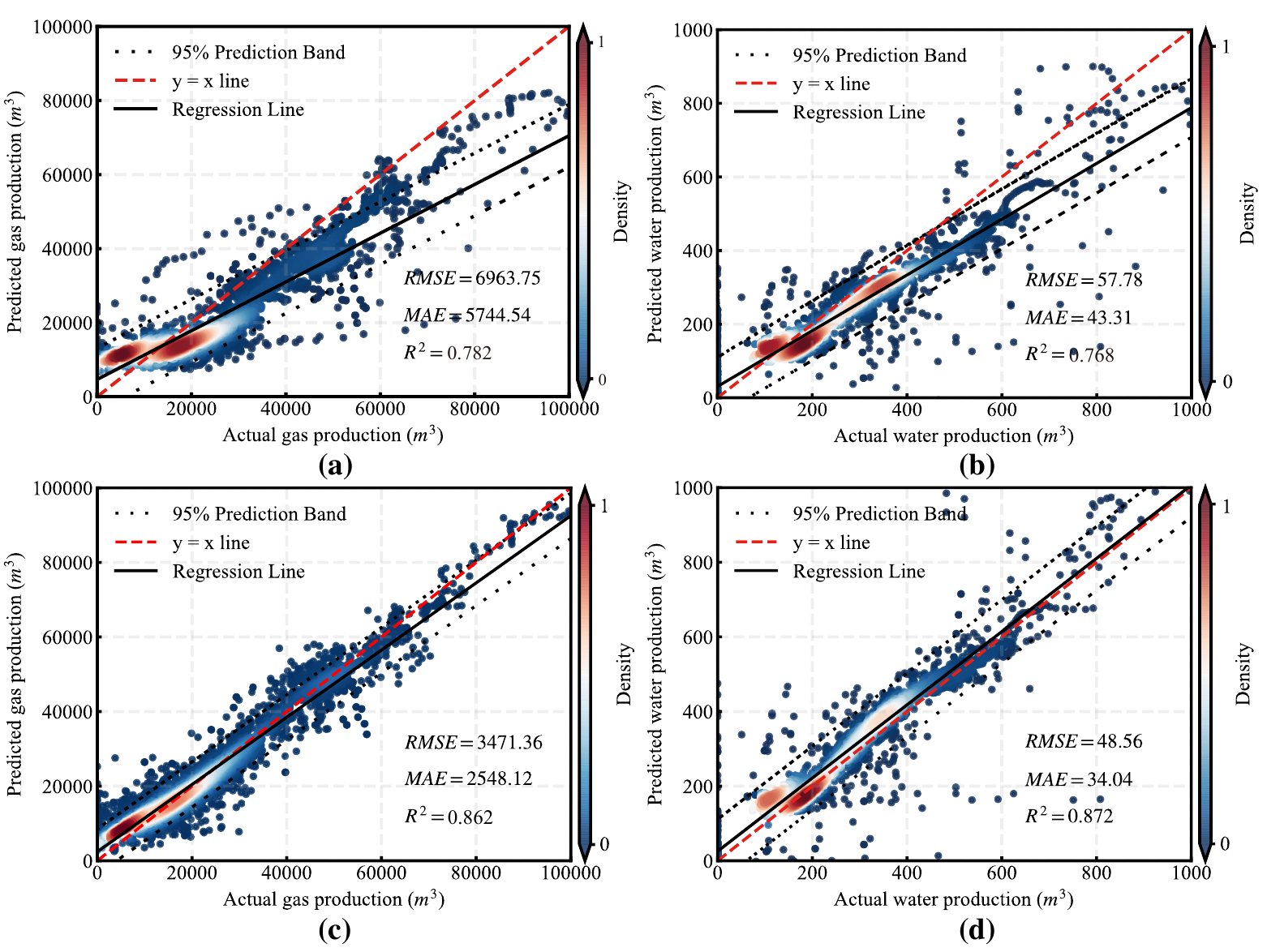}
    \caption{Scatter kernel density diagrams of different transfer learning methods on the target domain test set. Fig. 9 (a, b) illustrates the traditional transfer learning methodology for gas production and water production prediction, while Fig. 9 (c, d) showcases the proposed methodology result.}
    \label{fig:fig9}
\end{figure*}

First, we employ a greedy algorithm in the source domain to address the optimization problem outlined in \autoref{equ:equ4}. In this paper, we set the value of $K_0$ to 10, signifying that the time series is evenly divided into 10 segments, with each segment representing the smallest unit period that cannot be further subdivided. We then search for the optimal value of $K$ from the set \{2, 3, 4, 5, 6, 7, 8, 9, 10\}. To clarify the process, let's consider the case when $K$=2. We start by traversing through the nine candidate split points and calculate the split point $Q$ by maximizing the MMD distance between ${D}_{LQ}$ and ${D}_{QR}$. Once we determine the point $Q$, we proceed to consider $K$=3. Using the same approach, we determine the split point $Q$ and continue to identify the remaining segmentation points in a similar manner. In this specific example, the final value of $K$ is 6, resulting in the division of the entire source domain into 6 subsource domains denoted as $\mathcal{D}_{S_i}(i=1,2, \cdots, 6)$.

As shown in \autoref{fig:fig7}(b,d), the  traditional TL models exhibit a clear increase in predictive errors at certain time points, such as the 100th day and the 500th day. This decline occurs due to substantial fluctuations in gas and water production. However, in the initial phases of production, our proposed model, which combines domain adaptation and considers physical constraints, is capable of making accurate predictions about future output, even with limited production data, illustrated in \autoref{fig:fig7}(a,c).

Furthermore, in this study, we assess the effectiveness of different DL models by examining scatter density plots of actual values and predicted values. The red dashed line in \autoref{fig:fig9} represents the ideal state. Scatter points for traditional TL models exhibit significant deviations in predicting water production, indicating lower predictive accuracy. In contrast, scatter points for the model proposed in this study are mostly concentrated near the dashed line representing the ideal state and fall within a 95\% confidence interval. This phenomenon also confirms the stronger transfer capability of dynamic transfer for dynamic shale gas data.

Based on the results obtained from the test set, for the dynamic adaptive TL model, both root mean square error (RMSE) and mean absolute error (MAE) values are significantly lower than those of traditional TL models, while the coefficient of determination (R$^2$) is noticeably higher, reaching 0.862 and 0.872. For instance, in gas production prediction, the R$^2$ value is 9 percentage higher than the traditional TL model, while in water production prediction, it is 10 percentage points higher. This affirms that the model proposed in this study, through dynamic segmentation and forward transfer algorithms, partitions a single-source domain dataset into six subsource domains. As the number of subsource domains increases, they contribute complementarily to the target domain. The multisource model integrates this advantage to better represent the underlying representations of features, thereby enhancing its superiority in feature extraction and transfer for dynamic shale gas data compared to traditional TL models.

\section{Conclusion}
In this study, we propose a novel domain adaption transfer learning method with physical constrains to establish a reliable shale gas production prediction model under insufficient data. The effectiveness of the method is verified using the datasets from two blocks in southwestern China. The main conclusions are as follows:

(1) a dynamic segmentation method is employed to divided the shale gas source domain data into subsource domains based on their distributions. This enhances the contribution of valuable information from these subdomains to the target domain.

(2) Effective assurance of positive knowledge transfer is achieved by comparing the Maximum Mean Discrepancy (MMD) and the global average distance. This guarantees the transfer and retention of knowledge during the transfer learning process.

(3) By integrating transferable knowledge through domain adaptation, we can create a more comprehensive target domain model. This enhances the model's performance and adaptability by establishing a reliable and stable transfer learning path from the source domain to the target domain. Using data from the shale gas production base in southwest China, we have demonstrated that compared to the traditional fine-tuning method, the proposed approach increases the overall R$^2$ result by an average of 9.2\%.

(4) The proposed feature extractor incorporates physical constraints, leveraging geological and engineering information in shale gas production. This inclusion enhances the model's interpretability and adaptability to complex real-world environments, thereby increasing its practical applicability.

There are several limitations to the proposed methodology, which derives from two main aspects, namely, the data and the model. On the one hand, the data is the limited number of variables in the collected static shale gas dataset. Some factors are not included in the input, including maximum horizontal principal stress, vertical stress etc. Adding these factors could may improve the accuracy of predictions. On the other hand, we did not incorporate factors such as the measure of drainage in the shale gas production process into the model. Currently, determining the optimal time for these factors in shale gas production remains a key focus of research.
In the future, we plan to collect more real-world shale gas production datasets. These datasets will include more factors that impact shale gas production. With these datasets, we can utilize the proposed model to obtain higher-level representations of additional factors, enabling more effective and reliable predictions. To further investigate the impact of measures and other factors on production, we plan to explore the application of additional causal discovery algorithms in uncovering temporal causality. We have already gained some experience in causal discovery within production optimization for coalbed methane \cite{a53min2023interpretability53}.

\section*{Credit authorship contribution statement}
\textbf{Zhaozhong Yang:} Editing, Review, Conceptualization, Validation, Funding Acquisition. \textbf{Liangjie Gou}: Exit code, Formal analysis, Writing, Review, Editing. \textbf{Chao Min:} Review, Editing, Methodology. 
\textbf{Duo Yi}: Exit code, Visualization. \textbf{Xiaogang Li}: Methodology, Visualization, Funding acquisition. \textbf{Guoquan Wen:} Formal analysis, Writing, Review, Editing.

\section*{Declaration of competing interest}

The authors declare that they have no known competing financial interests or personal relationships that could have appeared to influence the work reported in this paper.

\section*{Data availability}

Data will be made available on request.

\section*{Acknowledgments}
This paper is supported by the International Cooperation Program of Chengdu City (No.2020-GH02-00023-HZ) and the Innovation Seedling Project of Sichuan Province "Research on prediction method of oil and gas field development index based on interpretable machine learning"(NO:2022034)

%% 设置参考文献排序方式和格式: Loading bibliography style file
%\bibliographystyle{cas-model2-names}
%\bibliographystyle{plain} % 按姓氏字母排列
\bibliographystyle{unsrt} % 按顺序排列

%% 设置加载的参考文献库: Loading bibliography database
\bibliography{cas-refs}
\end{document}